%% file: main.tex
\PassOptionsToPackage{table}{xcolor}
\documentclass[sigconf,9pt]{acmart}
\copyrightyear{2027}
\acmYear{2027}
\setcopyright{cc}
\setcctype{by}
\acmConference[ASPDAC '27]{32nd Asia and South Pacific Design Automation Conference}{January 25--28, 2027}{Tokyo, Japan}
\acmBooktitle{32nd Asia and South Pacific Design Automation Conference (ASPDAC '27), January 25--28, 2027, Tokyo, Japan}
\acmDOI{10.1145/3840404.3848213}
\acmISBN{979-8-4007-2923-2/2027/01}
\graphicspath{{./figs/}{./}}

\usepackage{graphicx}
\usepackage{microtype}
\usepackage{textcomp}
\usepackage{xcolor}
\usepackage{booktabs}
\usepackage{pifont}
\usepackage{pgfplots}
\usepackage{pgfplotstable}
\pgfplotsset{compat=1.8}
\usepackage{comment}
\usepackage{multirow}
\hypersetup{
    colorlinks = true,
    citecolor  = blue,
    linkcolor  = blue,
    urlcolor   = blue,
}
\usepackage{resizegather}

\usepackage[subrefformat=parens,farskip=0pt,justification=centering]{subfig}

\usepackage{url}
\usepackage{threeparttable}
\usepackage{physunits}
\usepackage[round-pad=false, round-mode=places,round-precision=2,group-separator={,},output-decimal-marker={.}]{siunitx}

\makeatletter
\newif\if@restonecol
\makeatother

\usepackage[linesnumbered,ruled,vlined]{algorithm2e}
\usepackage{algpseudocode}
\usepackage{amsmath,amsfonts}
\SetKwRepeat{Do}{do}{while}
\SetAlgoInsideSkip{small}
\allowdisplaybreaks

\usepackage{cleveref}
\Crefformat{figure}{Fig.~#2#1#3}                           %
\Crefname{subfigure}{Fig.}{Figs.}
\Crefname{figure}{Fig.}{Figs.}
\Crefformat{table}{TABLE~#2#1#3}                           %
\Crefname{table}{TABLE}{TABLEs}

\usepackage{tikz}
\usetikzlibrary{positioning,arrows.meta}
\usepackage{enumitem}
\usepackage{afterpage}

\newcommand{\minisection}[1]{\vspace{.06in}\noindent{\textbf{#1}}.}

\newcommand{\wns}{\textrm{WNS}}

\iftrue
    \setlist{leftmargin=12pt}
\fi

\begin{document}

\title{AgenticPD: A Stage-Aware Agentic Framework for Closed-Loop Physical Design Optimization}

\author{Shuo Ren}
\affiliation{\institution{The Chinese University of Hong Kong}\country{Hong Kong SAR}}
\email{sren@cse.cuhk.edu.hk}

\author{Zijin Cheng}
\affiliation{\institution{Southeast University}\city{Nanjing}\state{Jiangsu}\country{China}}
\email{220246802@seu.edu.cn}

\author{Yaohui Han}
\affiliation{\institution{The Chinese University of Hong Kong}\country{Hong Kong SAR}}
\email{yhhan25@cse.cuhk.edu.hk}

\author{Libo Shen}
\affiliation{\institution{The Chinese University of Hong Kong}\country{Hong Kong SAR}}
\email{lbshen24@cse.cuhk.edu.hk}

\author{Leilei Jin}
\affiliation{\institution{The Chinese University of Hong Kong}\country{Hong Kong SAR}}
\email{lljin@cse.cuhk.edu.hk}

\author{Wanting Tian}
\affiliation{\institution{The Chinese University of Hong Kong}\country{Hong Kong SAR}}
\email{wantingtian@cuhk.edu.hk}

\author{Rongliang Fu}
\affiliation{\institution{The Chinese University of Hong Kong}\country{Hong Kong SAR}}
\email{rongliangfu@outlook.com}

\author{Chao Wang}
\affiliation{\institution{Southeast University}\city{Nanjing}\state{Jiangsu}\country{China}}
\email{wangc@seu.edu.cn}

\author{Bei Yu}
\affiliation{\institution{The Chinese University of Hong Kong}\country{Hong Kong SAR}}
\email{byu@cse.cuhk.edu.hk}

\author{Tsung-Yi Ho}
\affiliation{\institution{The Chinese University of Hong Kong}\country{Hong Kong SAR}}
\email{tyho@cse.cuhk.edu.hk}

\renewcommand{\shortauthors}{Shuo Ren et al.}

\input{doc/0-abstract}

\maketitle
\pagestyle{plain}

\input{doc/1-intro}

\input{doc/2-prelim}
\input{doc/3-method}

\input{doc/4-exp}
\input{doc/5-conclusion}

\balance
\bibliographystyle{ACM-Reference-Format}
\bibliography{ref/reference}

\end{document}

%% file: doc/0-abstract.tex
\begin{abstract}
Physical design quality-of-results~(QoR) optimization is hard and expensive.
Choices made at one stage can help or hurt later stages.
Each evaluation requires a costly EDA run through the full flow.
While existing methods still treat optimization as flat parameter tuning or a LLM-based script generation task, we present AgenticPD\footnote{Our code is open-sourced at \url{https://github.com/Flians/AgenticPD}.\newline $^*$~Corresponding author: Rongliang Fu (rlfu@cse.cuhk.edu.hk).}, a stage-aware agentic framework for physical design QoR optimization.
Instead of re-running the full flow after every trial, AgenticPD is organized around the stage boundaries of the physical design flow, where a Judge Agent navigates the search and stage-specialized agents make local decisions within their own stage using stage-local tools.
Additionally, the agent harness in AgenticPD provides structured observations, execution history, and agent context management.
Experimental results show that AgenticPD achieves the best timing performance among the evaluated PD tuners while maintaining competitive power and area.
\end{abstract}

%% file: doc/1-intro.tex
\section{Introduction}
\label{sec:intro}
As modern chips grow in design complexity, achieving strong performance, power, and area~(PPA) under tight design schedules has become increasingly difficult~\cite{2024DAC_PD_opt}. Physical design (PD) plays a central role in this process~\cite{2022TCAD_PD_tuning_BO_beiyu}, since implementation decisions made during backend optimization directly affect final chip quality.
Moreover, physical design quality-of-results~(QoR) optimization is far from straightforward because the implementation flow spans multiple connected stages, including floorplan, placement, clock tree synthesis, and routing~\cite{2011_Andrew_PD_book}. Each stage has its own local objectives, optimization methods, and feedback metrics. Yet stage-local improvements do not always translate into better final QoR. A decision that appears highly effective at one stage may leave too little optimization space for later stages and can ultimately degrade final performance, power, or area. As a result, physical-design optimization must be understood not only at the level of individual stages, but also in terms of how decisions propagate across the full flow. Together, these cross-stage dependencies make QoR optimization a challenging full-flow problem in practical chip implementation~\cite{2011_Andrew_PD_book}.

\begin{figure}[t]
      \centering
      \includegraphics[width=\linewidth]{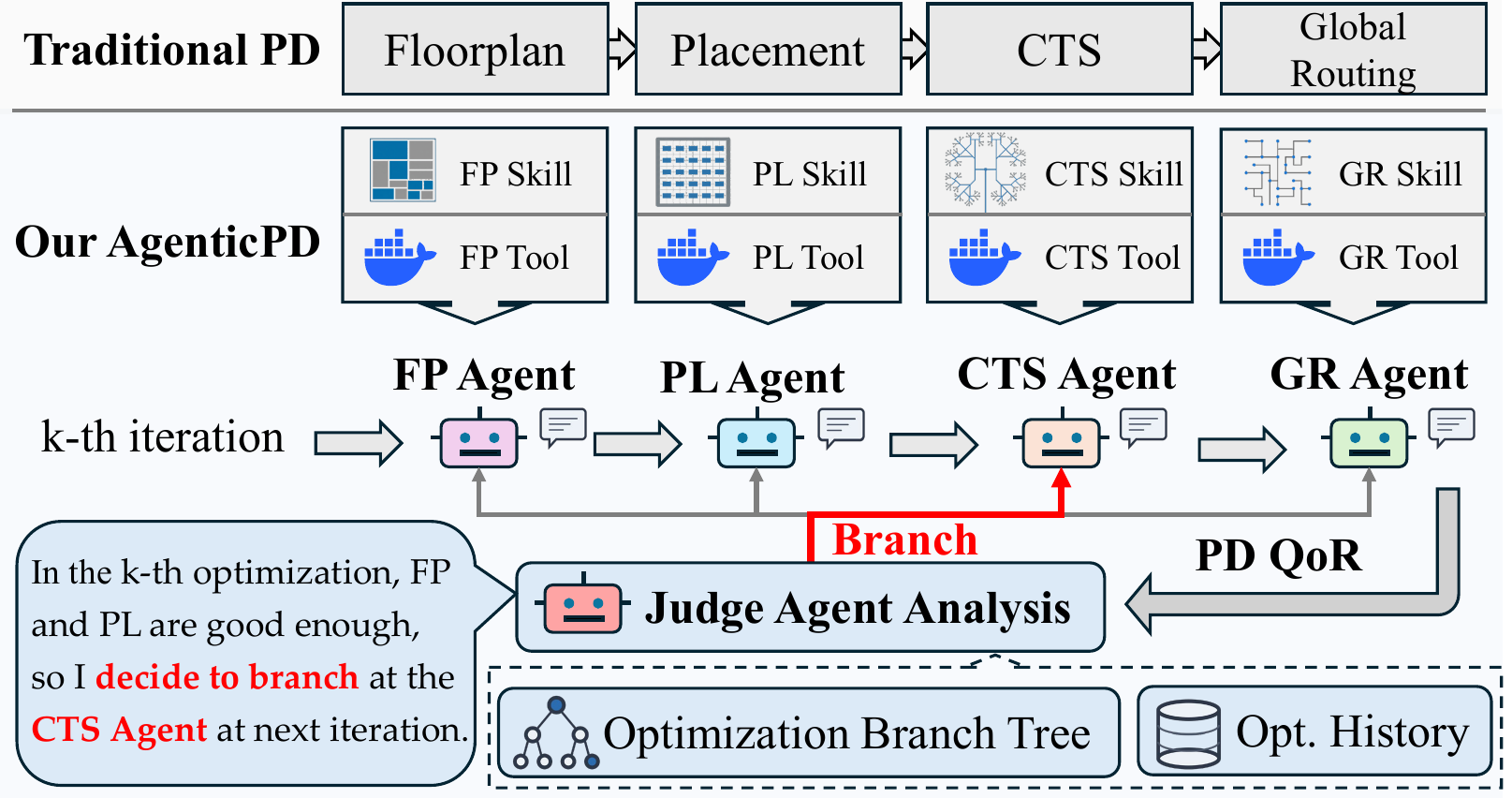}
      \caption{AgenticPD replaces black-box full-flow tuning with closed-loop, stage-aware multi-agent optimization.}
      \label{fig:intro-overview}
\end{figure}

Existing methods for physical design QoR optimization can be broadly divided into two categories.
The first category includes black-box tuning methods, such as AutoTuner~\cite{2021ICCAD_Andrew_METRICS2.1_AutoTuner}, PPATuner~\cite{2022DAC_PPATuner_beiyu}, REMOTune~\cite{2023TODAES_beiyu_REMOTune}, and FastTuner~\cite{2024ISPD_sklim_FastTuner}, which improve sampling efficiency by automatically exploring the configuration space.
However, these methods typically treat the physical design flow as a flat input-output mapping and do not explicitly exploit stage-wise structure, intermediate feedback, or reusable design states.

The second category includes recent LLM-based systems, such as ORFS-Agent~\cite{2025MLCAD_Andrew_ORFS-Agent}, OpenROAD-Assistant~\cite{2024MLCAD_OpenROAD_Assistant}, OpenROAD Agent~\cite{2025ICLAD_OpenROAD_Agent}, and ChatEDA~\cite{2024TCAD_byu_ChatEDA}, which rely on language models for script generation and document exploration.
However, these systems are still largely script-generation-like in how they organize optimization.
Such a formulation is not well suited to physical-design QoR optimization, because this task is multi-stage, state-dependent, and requires decisions to be revised according to real execution feedback.
Although these methods may incorporate feedback across trials, it remains at the full-flow configuration level; neither category closes the loop over reusable stage states and branch decisions.

To address these challenges, we present AgenticPD, a stage-aware agentic framework for physical design QoR optimization. As shown in~\Cref{fig:intro-overview}, AgenticPD organizes optimization around the structure of the full physical design flow and supports more effective exploration of the QoR search space, rather than treating each trial as an independent attempt.
This leads to stronger optimization outcomes than baselines. The main contributions are summarized as follows:
\begin{itemize}
      \item We formulate physical design QoR optimization as a stage-aware tree search that captures cross-stage dependencies and supports non-sequential exploration across the full physical design flow.
      \item We propose a hierarchical multi-agent framework that closes the loop between measured backend QoR and history-aware branch navigation, while specialized agents optimize individual physical design stages using stage-local feedback.
      \item We build an agent-ready physical design environment with stage-scoped actions, structured execution feedback, and stage-aware design-state reuse across explored branches.
      \item Experiments across six design and technology combinations show improved timing over the evaluated baselines while maintaining competitive power and area across both technologies.
\end{itemize}

%% file: doc/2-prelim.tex
\begin{figure*}[t!]
  \centering
  \includegraphics[width=\linewidth]{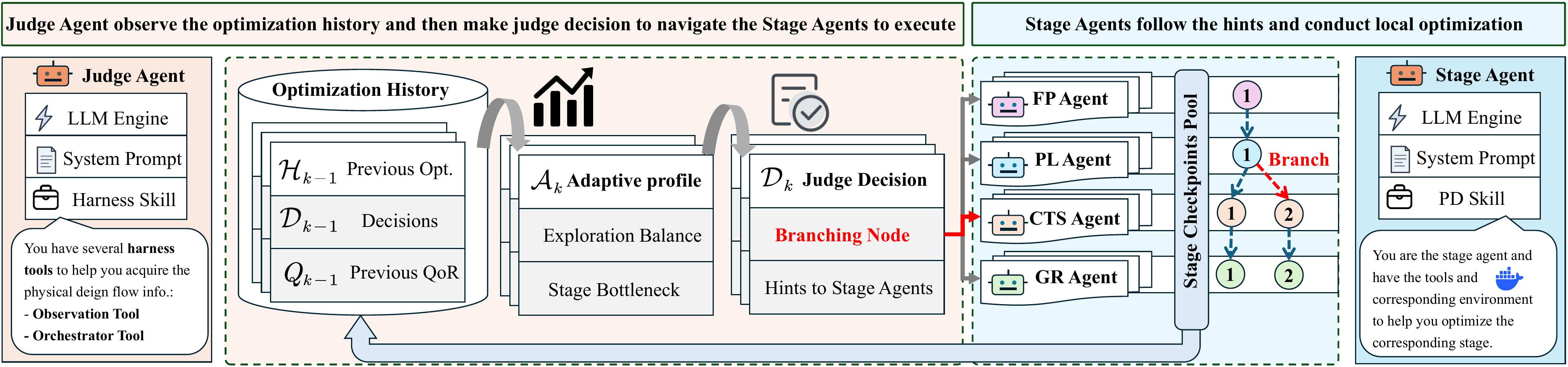}
  \caption{AgenticPD framework overview: measured QoR updates the optimization history, the Judge selects a branch node and stage, and downstream Stage Agents conditionally expand the selected subtree and return new feedback.}
  \label{fig:agent_system}
\end{figure*}

\section{Preliminaries and Motivations}
\label{sec:prelim}
\subsection{Physical Design}
In the physical design flow~\cite{2011_Andrew_PD_book}, we focus on four controllable stages:
floorplanning (FP), placement (PL), clock tree synthesis (CTS), and global routing (GR).
During each optimization run, candidates are evaluated after GR and only the selected candidate proceeds to detailed routing (DR). Headline comparisons use post-DR metrics; in-loop analyses are labeled in \Cref{sec:proxy}.
We formulate the physical design flow as an ordered sequence of stages:
\begin{equation}
  \mathcal{S} = (\mathrm{FP},\; \mathrm{PL},\; \mathrm{CTS},\; \mathrm{GR}),
  \label{eq:stage-sequence}
\end{equation}
where $s \in \mathcal{S}$ denotes a physical design stage.
We refer to the output of each stage as an \emph{intermediate design state}, which can be reused when alternative downstream decisions are explored.
Each stage $s$ has its own action space $\Theta_s$, representing the tunable parameters at that stage.
The full action space factorizes as
\begin{equation}
  \Theta_{\mathrm{PD}} \;=\; \Theta_{\mathrm{FP}} \times \Theta_{\mathrm{PL}}
  \times \Theta_{\mathrm{CTS}} \times \Theta_{\mathrm{GR}}.
  \label{eq:prelim_action_space}
\end{equation}
Because the stages execute in a fixed order, any selected stage~$b \in \mathcal{S}$ partitions the flow into two parts: the stages before~$b$, denoted $\mathrm{Bef}(b)$, and the stages after~$b$, denoted $\mathrm{Aft}(b)$.
For example, $\mathrm{Bef}(\mathrm{CTS}) = \{\mathrm{FP}, \mathrm{PL}\}$
and $\mathrm{Aft}(\mathrm{CTS}) = \{\mathrm{GR}\}$.
Moreover, a single complete flow execution is specified by an action tuple
\begin{equation}
  \mathbf{a} \;=\; \bigl(a(\mathrm{FP}),\; a(\mathrm{PL}),\;
  a(\mathrm{CTS}),\; a(\mathrm{GR})\bigr) \;\in\; \Theta_{\mathrm{PD}},
  \label{eq:action-tuple}
\end{equation}
where $a(s) \in \Theta_s$ denotes the action applied at stage~$s$.
Executing $\mathbf{a}$ runs the four stages in order and produces stage-level feedback $Q(s)$. We consider timing-primary physical design optimization: WNS and TNS are exposed as timing feedback, while post-GR WNS serves as the scalar criterion for incumbent selection; area and power are retained as secondary physical-quality metrics for final evaluation. Within the search budget~$N$, the optimizer seeks the candidate with the highest post-GR WNS:
\begin{equation}
  k^* = \arg\max_{0 \leq k < N} \mathrm{WNS}_k(\mathrm{GR}),
  \qquad \mathbf{a}^* = \mathbf{a}_{k^*}.
  \label{eq:timing-objective}
\end{equation}
During a standard optimization run, detailed routing is applied once to~$\mathbf{a}^*$ for final reporting.

\subsection{Agentic AI and Agentic EDA}
Recent progress in Agentic AI shows that language models can solve
complex tasks more effectively when they are allowed to reason, call
tools, inspect execution results, and revise decisions over multiple
steps rather than produce one static output~\cite{deepseek_v32_2025, qwen35_2026, kimi_k25_2026, Openai_chatgpt, Anthropic2024agents, han2026leafcell, fu2026mappingevolve}.
This interaction pattern is especially useful when task state evolves
during execution and when later actions depend on earlier observations.

The EDA community has also started to adopt this paradigm and apply agentic AI to EDA problems.
Cadence introduced ChipStack for AI-assisted chip design and verification~\cite{Cadence_ChipStack_2025}.
Synopsys introduced AgentEngineer for multi-agent automation of RTL and verification tasks \cite{Synopsys_AgentEngineer_2026}.
Siemens EDA launched Fuse EDA AI Agent for multi-tool automation from design to physical implementation \cite{Siemens_fuse_2026}.
However, these agentic EDA tools still mainly focus on design, verification, and workflow automation, while the use of agentic AI for physical design optimization remains underexplored.

\subsection{Motivations}
Physical-design QoR optimization is a natural fit for an agentic framework for three reasons.
(1)~It is not a one-shot prediction but a sequential process across FP, PL, CTS, and routing, where each stage has its own objectives, action space, and feedback yet depends on the state produced by earlier stages---closer to multi-step decision making than to flat parameter generation.
(2)~It requires both prior knowledge and execution feedback: prior knowledge guides stage-local choices at the start, but the true effect of a decision is revealed only after running the tools, so the optimizer must often revise or discard its initial judgment.
(3)~It requires repeated decision making and adjustment driven by execution feedback, which a one-shot script cannot naturally support.
\textbf{Taken together, these properties make physical-design QoR optimization a natural fit for an agentic, stage-aware framework.}

%% file: doc/3-method.tex
\section{Methodology}
\label{sec:method}

\input{doc/3.1-tree}
\input{doc/3.2-judge}
\input{doc/3.3-stage}
\input{algs/alg-tree-search}
\input{tables/tab-designs}
\input{tables/tab-ablation-single}
\input{doc/3.5-workflow}

%% file: doc/3.1-tree.tex
\subsection{Stage-Aware Optimization Tree}
\label{subsec:tree}

Unlike trial-based tuners that generate an independent full-flow configuration at each iteration, AgenticPD can revisit a retained state, select a physical design stage to revise, and expand only the corresponding downstream subtree.

We represent this search as a rooted tree~$\mathcal{T}$.
The root~$n_0$ represents the post-synthesis design~$D$ before any PD stage runs.
Each time a stage~$s$ is executed in iteration~$k$, a node is created:
\begin{equation}
  n_k^s \;=\; \bigl(x_k(s^+),\; a_k(s),\; Q_k(s)\bigr),
  \label{eq:node-def}
\end{equation}
where $x_k(s^+)$ is the resulting intermediate design state, $a_k(s) \in \Theta_s$ is the applied action, and $Q_k(s)$ is the observed stage-level QoR.
Every root-to-leaf path traverses all four stages in order and corresponds to one complete PD flow with action tuple~$\mathbf{a}_k$ (\Cref{eq:action-tuple}).

\minisection{Branching}
Not every iteration needs to start from scratch.
If a previous iteration already produced a good result at some early stage, re-running that stage with a different action is wasteful; the optimizer should focus its budget on the stages that still have room to improve.
Branching makes this possible.
At iteration~$k$, the optimizer selects a retained node~$\hat{n}$ whose stored state is the input to a branching stage~$b \in \mathcal{S}$ and starts a new branch from there.
All stages before~$b$ are inherited from the ancestor path leading to~$\hat{n}$---the $\mathrm{Bef}(b)$ results are reused at zero cost---and only stage~$b$ itself and the stages in $\mathrm{Aft}(b)$ are re-executed with new actions.
Let $q=|\mathcal{S}|$ and index the ordered stages by $1,\ldots,q$. Their causal order is encoded by~$C\in\{0,1\}^{q\times q}$, and the branch choice activates a downstream suffix through
\begin{subequations}
\label{eq:branch-expansion}
\begin{align}
  C_{ij}&=\mathbb{I}[i\leq j], \qquad
  \mathbf{m}_k=C^\top\mathbf{e}(b_k),
  \label{eq:stage-activation}\\
  \mathcal{T}_{k}&=\mathcal{T}_{k-1}\cup
  \{n_k^{s_j}:m_{k,j}=1\}.
  \label{eq:tree-grow}
\end{align}
\end{subequations}
where $\mathbf{e}(b_k)$ is the one-hot vector of the branching stage. The first relation maps the Judge's branch choice to stage activation: $m_{k,j}=0$ marks a reusable upstream stage and $m_{k,j}=1$ activates a downstream stage. The second relation attaches exactly those activated nodes to the tree. Together, they define the branch-conditioned subtree expansion while preserving the upstream design state stored at~$\hat{n}$.
When $b_k = \mathrm{FP}$ (the first stage), branching from the root reduces to running a completely new flow from scratch.

This tree structure turns flat independent trials into a structured search. It retains every explored design state, not only the current best candidate, so the optimizer can return to another promising node when refinement along one branch reaches a local limit.
At each iteration, the Judge observes the search history, selects a branch node and branching stage, and generates stage-specific hints. The downstream Stage Agents then execute the remaining stages from the selected design state. The resulting nodes are added to the tree, while the best-WNS incumbent is updated independently.

%% file: doc/3.2-judge.tex
\subsection{History-Aware Branch Navigation}
\label{subsec:judge_agent}
As shown in \Cref{fig:agent_system}, the Judge serves as a navigation policy over the optimization tree rather than a generator of full-flow parameters. We instantiate this policy using a general-purpose LLM~$\mathcal{L}$, a role-specific system prompt~$\mathcal{P}_J$, and the Agent Harness~$\mathcal{U}_J$. The prompt defines its branching output and search guidance, while the harness supplies the structured observation and executes the resulting branch decision (\Cref{subsec:harness}).
At iteration~$k$, the optimization history~$\mathcal{H}_k$ records the selected node, branching stage, and downstream QoR from every previous iteration. Rather than directly generating a full-flow parameter vector, the Judge uses this history to decide where to resume the search and which stages to revise. The Observation Tool first computes an \textit{adaptive profile}
\begin{equation}
  \mathcal{A}_k \;=\; \bigl(\{E(n)\}_{n \in \mathcal{T}_k},\; \{B(s)\}_{s \in \mathcal{S}}\bigr),
  \label{eq:adaptive-profile}
\end{equation}
where the exploration balance~$E(n)$ measures how often node~$n$ has been revisited and the stage bottleneck~$B(s)$ measures how far stage~$s$ falls behind its best-known result. Together with a compressed tree snapshot, $\mathcal{A}_k$ forms the Judge's search state. The Judge then produces
\begin{equation}
  \mathcal{D}_k
  \;=\; (\hat{n}_k,\; b_k,\; h_k)
  \;=\; \mathrm{Judge}(\mathcal{H}_k,\mathcal{A}_k),
  \label{eq:judge-decision}
\end{equation}
where $\hat{n}_k$ is the branch node, $b_k \in \mathcal{S}$ is the branching stage, and $h_k$ contains one hint for each downstream Stage Agent. This structured decision jointly determines where to resume the search, which stage to revise, and how to guide downstream expansion. Guided by $B(s)$ and $E(n)$, the Judge balances bottleneck repair with exploration of less-visited nodes. The selected node need not be the current incumbent; any retained node can seed a promising downstream branch. After execution, the new branch and its stage-level results are appended to $\mathcal{T}_k$ and $\mathcal{H}_k$, while the best-WNS incumbent is updated independently according to \Cref{eq:timing-objective}.

%% file: doc/3.3-stage.tex
\begin{figure}[t!]
  \centering
  \includegraphics[width=0.95\linewidth]{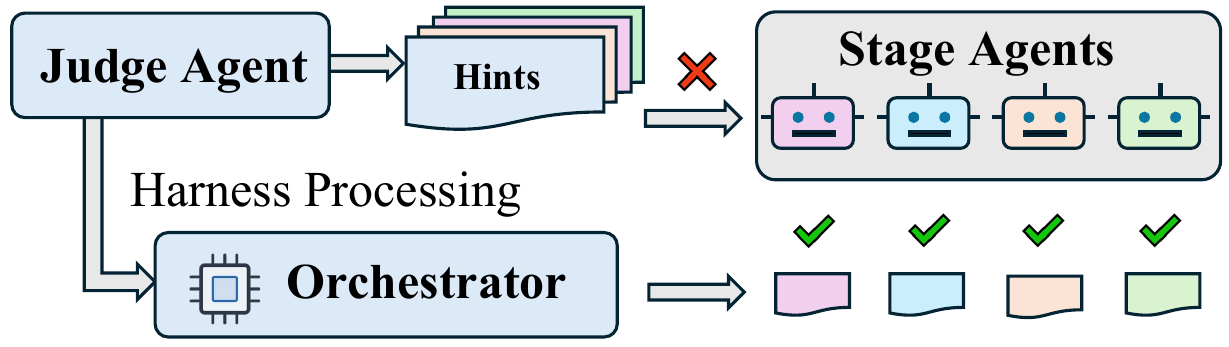}
  \caption{Agent Harness: the Judge's hints are dispatched to downstream Stage Agents, each of which executes through its own PD Skill.}
  \label{fig:orchestrator}
\end{figure}

\subsection{Branch-Conditioned Agent Execution}
\label{subsec:stage_agent}
\label{subsec:harness}
Given the Judge decision~$\mathcal{D}_k=(\hat{n}_k,b_k,h_k)$, AgenticPD reuses the design state at~$\hat{n}_k$ and activates only the Stage Agents selected by~$\mathbf{m}_k$ in \Cref{eq:stage-activation}. For each activated stage~$s$, the harness assembles
\begin{equation}
  ctx_k(s) \;=\; (\{Q_k(i)\}_{i \in \mathrm{Bef}(s)},\; e_s,\; K_s,\; \mathrm{hint}_s),
  \label{eq:stage-context}
\end{equation}
where the four terms are the upstream QoR on the current branch, a cross-iteration experience summary, a retrieved knowledge block, and the Judge's stage-specific guidance, respectively. GraphRAG retrieves $K_s=\mathrm{Retrieve}(s,ctx_k(s))$ from a documentation-grounded graph. The graph links documented stage-level actions to downstream metrics; its text blocks are embedded offline and matched to the current stage condition.

Let $d_j$ be the number of tunable parameters at stage~$s_j$, stack the stage-local proposals as~$\widetilde{\mathbf{a}}_k$, and let~$\mathbf{a}_{\hat{n}_k}$ denote the configuration on the retained path. The branch-conditioned expansion is
\begin{equation}
  \begin{aligned}
    M_k
      &= \operatorname{blkdiag}
      \left(m_{k,1}I_{d_1},\ldots,m_{k,q}I_{d_q}\right), \\
    \mathbf{a}_k
      &= (I-M_k)\mathbf{a}_{\hat{n}_k}
      +M_k\widetilde{\mathbf{a}}_k.
  \end{aligned}
  \label{eq:conditional-expansion}
\end{equation}
Here, $I-M_k$ preserves parameters on the retained path, whereas $M_k$ replaces the activated blocks.
Starting from~$x(\hat{n}_k)$, each activated agent proposes $\widetilde{a}_k(s)=\pi_s(ctx_k(s))$, and the backend advances the design through $x_k(s^+)=F_s(x_k(s),\widetilde{a}_k(s))$, passing~$Q_k(s)$ downstream. The Agent Harness realizes this execution through three functions. (1)~The \textit{Observation Tool} reads~$\mathcal{T}_k$ and~$\mathcal{H}_k$, computes~$\mathcal{A}_k$ (\Cref{eq:adaptive-profile}), and constructs a compressed tree snapshot for the Judge. (2)~The \textit{Orchestration Tool} converts~$\mathcal{D}_k$ into the masked configuration in \Cref{eq:conditional-expansion}, invokes the activated agents in pipeline order, and dispatches their stage-specific hints (\Cref{fig:orchestrator}). (3)~Each activated agent combines the shared LLM~$\mathcal{L}$ with a stage-specific prompt~$\mathcal{P}_s$ and \textit{PD Skill}~$\mathcal{U}_s$.

%% file: algs/alg-tree-search.tex
\begin{algorithm}[t]
    \caption{Stage-Aware Branch Search in AgenticPD}
    \label{alg:workflow}
    \small
    \setlength{\hsize}{0.95\linewidth}
    \KwIn{Design $D$, total evaluation budget $N$, initial action tuple $\mathbf{a}_0$}
    \KwOut{Best action tuple $\mathbf{a}^{*}$ and its post-DR QoR $Q^{\mathrm{DR}}$}
    Run $\mathbf{a}_0$ through GR and observe $\{Q_0(s)\}$ \tcp*{\Cref{eq:action-tuple}}
    Init tree $\mathcal{T}$ with results\;
    $(\mathbf{a}^{*}, Q^{*}) = (\mathbf{a}_0, Q_0(\mathrm{GR}))$\;
    $\mathcal{H} = \emptyset$\;
    \For{\textup{$k = 1$ to $N-1$}}{
        $\mathcal{A}_k = \textsc{ObservationTool}(\mathcal{T},\, \mathcal{H})$ \tcp*{adaptive profile from~\Cref{eq:adaptive-profile}}
        $(\hat{n}_k,\, b_k,\, h_k) = \textsc{Judge}(\mathcal{H},\, \mathcal{A}_k)$ \tcp*{decision from~\Cref{eq:judge-decision}}
        Reuse the state after $\mathrm{Bef}(b_k)$ from node $\hat{n}_k$ \;
        \For{\textup{each stage $s \in \{b_k\} \cup \mathrm{Aft}(b_k)$}}{
            $ctx_k(s) = \textsc{BuildContext}(s,\; \hat{n}_k,\; \mathrm{hint}_s)$ \tcp*{\Cref{eq:stage-context}}
            $a_k(s) = \textsc{StageAgent}_s(ctx_k(s))$ \tcp*{\Cref{subsec:stage_agent}}
            $Q_k(s) = \textsc{ExecuteStage}(s,\;
            a_k(s))$ \tcp*{\Cref{subsec:stage_agent}}
        }
        Update $\mathcal{T}$ and $\mathcal{H}$ \tcp*{\Cref{eq:tree-grow}}
        Update $(\mathbf{a}^{*},\, Q^{*})$ if $Q_k(\mathrm{GR})$ has higher WNS \tcp*{\Cref{eq:timing-objective}}
    }
    $Q^{\mathrm{DR}} = \textsc{DetailedRoute}(\mathbf{a}^{*})$; report $(\mathbf{a}^{*}, Q^{\mathrm{DR}})$\;
\end{algorithm}

%% file: tables/tab-designs.tex
\begin{table}[t!]
\centering
\caption{Benchmark designs evaluated on two process nodes.}
\label{tab:designs}
\scriptsize
\setlength{\tabcolsep}{4pt}
\begin{tabular}{@{}clrrlrr@{}}
\toprule
\textbf{Function} & \textbf{Design} & \textbf{Cells} & \textbf{Clk} &
\textbf{Design} & \textbf{Cells} & \textbf{Clk} \\
\midrule
AES cipher   & AES-128/SKY130HD & 12{,}787 & 4.50 ns  & AES-128/ASAP7 & 15{,}621 & 0.40 ns \\
RISC-V core  & ibex/SKY130HD    & 15{,}945 & 10.00 ns & ibex/ASAP7    & 18{,}340 & 1.26 ns \\
JPEG encoder & JPEG/SKY130HD    & 38{,}216 & 5.50 ns  & JPEG/ASAP7    & 54{,}354 & 0.90 ns \\
\bottomrule
\end{tabular}
\end{table}

%% file: tables/tab-ablation-single.tex
\begin{table*}[t]
\centering
\caption{AgenticPD vs.\ Vanilla LLM on ASAP7.
Bold\,=\,AgenticPD is the best.}
\label{tab:ablation-single}
\footnotesize
\begin{threeparttable}
\begin{tabular}{ll rrrr rrrr rrrr}
\toprule
& & \multicolumn{4}{c}{\textbf{Qwen3.5~\cite{qwen35_2026}}}
  & \multicolumn{4}{c}{\textbf{Kimi-K2.5~\cite{kimi_k25_2026}}}
  & \multicolumn{4}{c}{\textbf{DeepSeek-V3.2~\cite{deepseek_v32_2025}}} \\
\cmidrule(lr){3-6} \cmidrule(lr){7-10} \cmidrule(lr){11-14}
\textbf{Method} & \textbf{Design}
 & WNS$\uparrow$ & TNS$\uparrow$ & Area$\downarrow$ & Power$\downarrow$
 & WNS$\uparrow$ & TNS$\uparrow$ & Area$\downarrow$ & Power$\downarrow$
 & WNS$\uparrow$ & TNS$\uparrow$ & Area$\downarrow$ & Power$\downarrow$ \\
\midrule
\multirow{3}{*}{Vanilla LLM}
 & AES  & $-$50.749 & $-$2958.5 & 2.1 & 0.159
        & $-$39.768 & $-$2085.8 & 2.1 & 0.159
        & $-$38.338 & $-$2278.2 & 2.1 & 0.158 \\
 & ibex & $-$66.140 & $-$3325.2 & 2.7 & 0.048
        & $-$66.140 & $-$3325.2 & 2.7 & 0.048
        & $-$66.140 & $-$3325.2 & 2.7 & 0.048 \\
 & JPEG & 10.646 & 0 & 6.8 & 0.095
        & $-$5.006 & $-$7.5 & 6.7 & 0.092
        & $-$5.006 & $-$7.5 & 6.7 & 0.092 \\
\midrule
\multirow{3}{*}{AgenticPD}
 & AES  & \textbf{$-$35.457} & \textbf{$-$2654.4} & \textbf{2.0} & \textbf{0.157}
        & $-$40.064 & $-$2561.0 & \textbf{2.0} & \textbf{0.158}
        & \textbf{$-$32.271} & \textbf{$-$1811.3} & \textbf{2.0} & \textbf{0.157} \\
 & ibex & $-$99.135 & $-$11271.5 & \textbf{2.6} & \textbf{0.046}
        & \textbf{$-$65.416} & \textbf{$-$1837.8} & 2.7 & 0.048
        & \textbf{$-$53.867} & \textbf{$-$1098.7} & 2.7 & 0.048 \\
 & JPEG & \textbf{55.915} & 0 & \textbf{6.7} & \textbf{0.092}
        & \textbf{56.711} & \textbf{0} & 6.7 & 0.092
        & \textbf{50.569} & \textbf{0} & 6.7 & 0.092 \\
\midrule
\multicolumn{2}{l}{\textit{Geomean impr.}}
 & \multicolumn{12}{c}{Geomean gains: Performance 2.4\%,\; Area 2.2\%,\; Power 1.1\%} \\
\bottomrule
\end{tabular}
\begin{tablenotes}[flushleft]
\footnotesize
\item[*] WNS/TNS are in ps ($\uparrow$); area is in $10^3\,\mu\text{m}^2$ and power in W ($\downarrow$). Percentage improvements are computed as one minus the geometric mean of per-case AgenticPD\,/\,Vanilla LLM ratios, expressed as percentages, with performance measured by effective clock period ($\mathrm{ECP}=T_{\mathrm{clk}}-\mathrm{WNS}$) for consistent cross-design comparison as in~\cite{2025MLCAD_Andrew_ORFS-Agent}.
\end{tablenotes}
\end{threeparttable}
\end{table*}

%% file: doc/3.5-workflow.tex
\subsection{Overall AgenticPD Algorithm}
\label{subsec:workflow}
Algorithm~\ref{alg:workflow} integrates history-aware branch navigation, intermediate-state reuse, and conditional downstream expansion.

Lines~1--4 initialize the search by running $\mathbf{a}_0$ through GR and recording the results in the tree~$\mathcal{T}$ and history~$\mathcal{H}$.
Lines~6--14 describe one outer-loop iteration.
The Observation Tool first constructs the adaptive profile~$\mathcal{A}_k$ from the current tree and history (Line~6).
The Judge then produces the branching decision~$\mathcal{D}_k$ (Line~7).
The system reuses $\mathrm{Bef}(b_k)$ from node~$\hat{n}_k$ (Line~8) and re-executes $\{b_k\} \cup \mathrm{Aft}(b_k)$ (Lines~9--12): for each stage~$s$, the harness builds $ctx_k(s)$, and the Stage Agent decides $a_k(s)$ and executes the stage.
Lines~13--14 update the tree, history, and best-WNS incumbent.
Line~15 applies detailed routing to the final incumbent and reports its post-DR QoR.

\input{figs/fig_grt_proxy_pie}

Tree exploration and incumbent selection are decoupled. The tree~$\mathcal{T}$ retains every completed node for future branching, whereas the incumbent~$\mathbf{a}_k^*$ records the highest-post-GR-WNS candidate observed through iteration~$k$, following \Cref{eq:timing-objective}. The Judge may therefore explore a temporarily worse branch without discarding previously discovered candidates. This workflow brings together the stage-specific expertise of the Stage Agents and the decision-making capabilities of the Judge Agent.

%% file: figs/fig_grt_proxy_pie.tex
\definecolor{clrFP}{HTML}{D0D0D0}      %
\definecolor{clrPL}{HTML}{B0B0B0}      %
\definecolor{clrCTS}{HTML}{909090}     %
\definecolor{clrGR}{HTML}{E8915A}      %
\definecolor{clrDR}{HTML}{5B9BD5}      %
\begin{figure}[t!]
  \centering
  \begin{minipage}[c]{0.035\linewidth}
    \centering
    \rotatebox{90}{\scriptsize Normalized WNS}
  \end{minipage}%
  \begin{minipage}[c]{0.955\linewidth}
    \includegraphics[width=\linewidth]{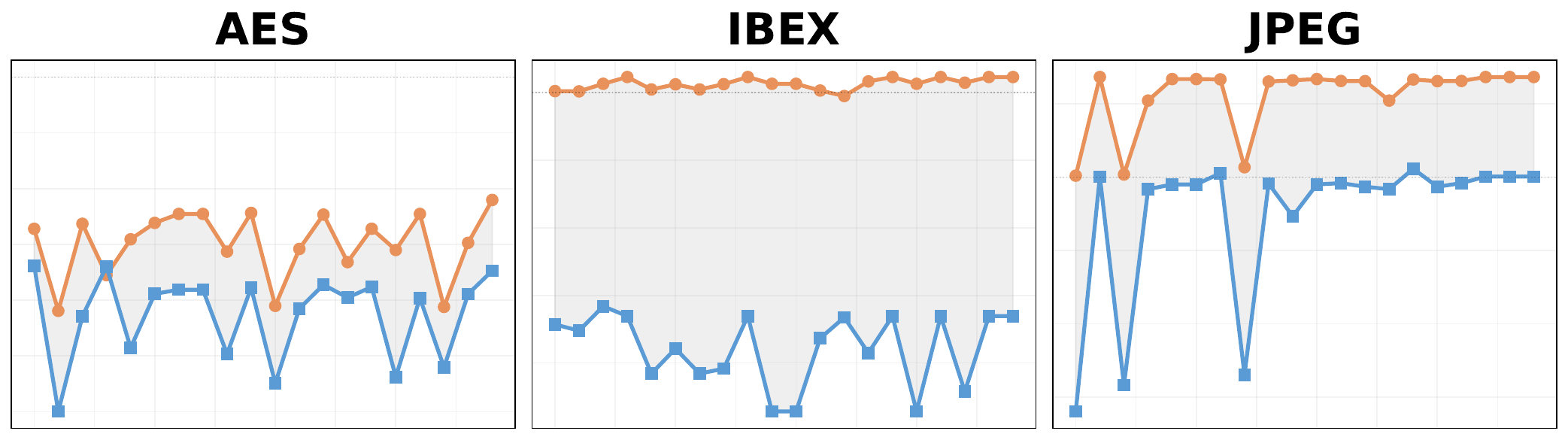}
  \end{minipage}
  {\scriptsize Sampled Configuration\par}

  \vspace{2pt}
  \makebox[\linewidth][c]{\hspace*{-0.16cm}%
  \begin{tikzpicture}
    \def\barw{6.85}
    \def\barh{0.42}
    \node[anchor=east, font=\scriptsize] at (-0.37,{0.5*\barh})
      {Time distribution};
    \begin{scope}[xshift=-0.25cm]
    \fill[clrFP, fill opacity=0.85]
      (0,0) rectangle ({0.11*\barw},\barh);
    \fill[clrPL, fill opacity=0.85]
      ({0.11*\barw},0) rectangle ({0.22*\barw},\barh);
    \fill[clrCTS, fill opacity=0.85]
      ({0.22*\barw},0) rectangle ({0.33*\barw},\barh);
    \fill[clrGR, fill opacity=0.85]
      ({0.33*\barw},0) rectangle ({0.539*\barw},\barh);
    \fill[clrDR, fill opacity=0.85]
      ({0.539*\barw},0) rectangle (\barw,\barh);
    \draw[gray!60, line width=0.3pt] (0,0) rectangle (\barw,\barh);
    \draw[white, line width=0.5pt] ({0.33*\barw},0) -- ({0.33*\barw},\barh);
    \draw[white, line width=0.5pt] ({0.539*\barw},0) -- ({0.539*\barw},\barh);
    \node[font=\scriptsize\bfseries, text=black]
      at ({0.165*\barw},{0.5*\barh}) {FP/PL/CTS 21.7\%};
    \node[font=\scriptsize\bfseries]
      at ({0.4345*\barw},{0.5*\barh}) {GR 27.2\%};
    \node[font=\scriptsize\bfseries]
      at ({0.7695*\barw},{0.5*\barh}) {DR 51.1\%};
    \end{scope}
  \end{tikzpicture}
  }

  \vspace{3pt}
  \begin{tikzpicture}
    \fill[black, fill opacity=0.06, rounded corners=2pt]
      (-0.15,-0.12) rectangle (8.3,0.42);
    \draw[clrGR, line width=1.5pt] (0,0.15) -- (0.4,0.15);
    \fill[clrGR] (0.2,0.15) circle (2pt);
    \node[right, font=\scriptsize] at (0.45,0.15) {In-loop};
    \draw[clrDR, line width=1.5pt] (2.0,0.15) -- (2.4,0.15);
    \fill[clrDR] (2.1,0.05) rectangle (2.3,0.25);
    \node[right, font=\scriptsize] at (2.45,0.15) {Post-route};
    \fill[clrFP, fill opacity=0.85]  (4.0,0) rectangle (4.2,0.3);
    \fill[clrPL, fill opacity=0.85]  (4.2,0) rectangle (4.4,0.3);
    \fill[clrCTS, fill opacity=0.85] (4.4,0) rectangle (4.6,0.3);
    \draw[gray!70, line width=0.3pt] (4.2,0) -- (4.2,0.3);
    \draw[gray!70, line width=0.3pt] (4.4,0) -- (4.4,0.3);
    \node[right, font=\scriptsize] at (4.65,0.15) {FP/PL/CTS};
    \fill[clrGR, fill opacity=0.85] (6.2,0) rectangle (6.5,0.3);
    \node[right, font=\scriptsize] at (6.55,0.15) {GR};
    \fill[clrDR, fill opacity=0.85] (7.4,0) rectangle (7.7,0.3);
    \node[right, font=\scriptsize] at (7.75,0.15) {DR};
  \end{tikzpicture}

\caption{Consistency between in-loop and post-route WNS across sampled configurations, with runtime distribution by stage on ASAP7.}
  \label{fig:grt_proxy}
\end{figure}

%% file: doc/4-exp.tex
\section{Experiments}
\label{sec:exp}
\subsection{Experimental Setup}
\label{sec:setup}
Experiments run on Ubuntu 22.04 LTS with dual Intel Xeon Gold 6426Y processors and 256 GB RAM.
We evaluate AES~\cite{design_aes}, ibex~\cite{design_ibex}, and JPEG~\cite{design_jpeg} on the SkyWater 130\,nm HD library~\cite{technode_sky130pdk} and ASAP 7\,nm predictive PDK~\cite{technode_ASAP7}, using a 20-evaluation backend budget. \Cref{tab:designs} summarizes their post-synthesis sizes and target clock constraints.
Post-GR WNS serves as the scalar incumbent-selection criterion, whereas complete post-route QoR is used for final evaluation. TNS reports aggregate timing violations, effective frequency is derived from the target clock period and post-route WNS, and area and power report non-timing physical quality. All optimization methods share the same budget.
All methods use the same backend~\cite{OpenROAD,openroadflow}, action space, and stage implementations.\footnote{Our public repository implements AgenticPD's history-aware Judge orchestration, stage-specialized agents, tree-structured branch navigation, and reusable intermediate design states, together with experiment configurations, session inspection tools, and mock-mode verification instructions at \url{https://github.com/Flians/AgenticPD}.}
\begin{figure}[t]
  \centering
  \vspace{-0.05em}
  \includegraphics[width=\linewidth]{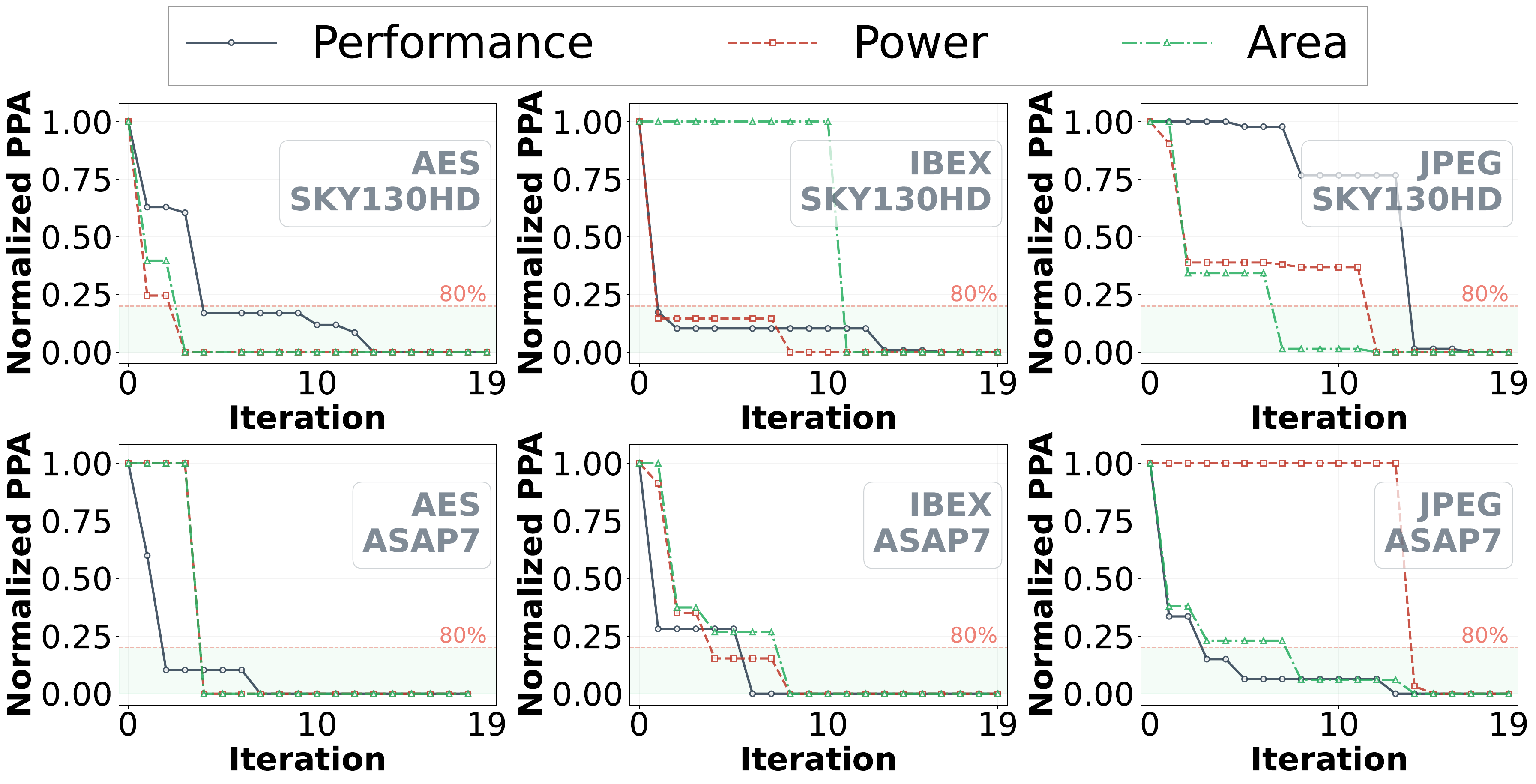}
  \caption{Normalized PPA convergence curves.}
  \label{fig:ppa-conv}
\end{figure}

\subsection{AgenticPD Mechanisms and Ablations}
\label{sec:proxy}
We examine AgenticPD from five complementary perspectives: in-loop QoR fidelity, robustness across LLM backbones, optimization dynamics, branching mechanisms, and the contribution of its Judge.
\noindent \textbf{QoR fidelity.} AgenticPD ranks candidate configurations by the in-loop WNS signal, while headline comparisons report post-route QoR. To verify its signoff fidelity, we re-run detailed routing and signoff extraction for every configuration explored by AgenticPD on SKY130HD. In-loop and post-route WNS agree closely: Spearman $\rho = 0.86$ (AES), $0.83$ (ibex), $0.93$ (JPEG), and Pearson $r = 0.96$ / $0.96$ / $0.91$. On ASAP7, \Cref{fig:grt_proxy} shows corresponding trends across sampled configurations.
Detailed routing takes a median of 38--75 minutes per configuration on identical hardware, so carrying a 20-candidate optimization through DR would add 13--25 hours per session. The in-loop signal removes this dominant cost from every iteration while preserving the ranking that guides optimization.

\input{figs/fig_branch_stages}

\noindent\textbf{Architecture robustness across LLM backbones.} To evaluate whether the agentic architecture itself drives the improvement, we compare AgenticPD against a Vanilla LLM baseline, a flat single agent that shares the same agent-friendly environment integration as AgenticPD. It uses the same 20-evaluation budget and Docker backend, proposing one complete parameter configuration per iteration while observing the full evaluation history.

We compare the two systems on ASAP7 with DeepSeek-V3.2~\cite{deepseek_v32_2025}, Qwen3.5~\cite{qwen35_2026}, and Kimi-K2.5~\cite{kimi_k25_2026}.
\Cref{tab:ablation-single} reports post-route PPA on three ASAP7 benchmarks. AgenticPD wins on WNS in 7 of 9 pairs. The gains are largest on hard cases: on JPEG, AgenticPD achieves timing closure across all three backbones, while Vanilla LLM fails or barely closes. Winning pairs span all three backbones, so the benefit is not confined to one model. DeepSeek-V3.2 delivers the best overall timing, so we adopt it as the default when an experiment requires an LLM engine.

\input{tables/tab-qor-final}

\noindent\textbf{Optimization dynamics.} With DeepSeek-V3.2, we analyze one optimization run on each of the six benchmarks.

\Cref{fig:ppa-conv} plots best-so-far ECP~\cite{2025MLCAD_Andrew_ORFS-Agent}, power, and area normalized between their initial and best observed values. Across all benchmarks, these metrics realize at least 80\% of their total improvement by iterations 8--14, indicating that useful configurations emerge well before the budget is exhausted.
At $N{=}20$, the compressed Judge profile and stage-local contexts keep the peak prompt across all agents and benchmarks to 43K tokens, within the context windows of all evaluated LLMs~\cite{deepseek_v32_2025,kimi_k25_2026,qwen35_2026}.

\noindent\textbf{Branching mechanisms.} \Cref{fig:branch-stages} shows the distribution of branching stages selected by the Judge across the six optimization runs analyzed above.
The selections are balanced across FP (28.1\%), CTS (27.2\%), GR (23.7\%), and PL (21.1\%), showing that the Judge does not collapse to a single restart stage. Early-stage branches rebuild the downstream flow, whereas later branches preserve proven upstream states and target localized timing or routing bottlenecks.

Combining the branch distribution in \Cref{fig:branch-stages} with the stage runtime profile in \Cref{fig:grt_proxy}, about 72\% of decisions select $b_k \in \{\mathrm{PL},\mathrm{CTS},\mathrm{GR}\}$, reuse the cached stages in $\mathrm{Bef}(b_k)$, and execute only $\{b_k\}\cup\mathrm{Aft}(b_k)$. These avoided upstream executions indicate backend-time savings over full-flow baselines without additional EDA evaluations.

\input{tables/tab-stability}
\input{tables/tab-judge}

\noindent\textbf{Stability and Judge ablation.} Because the backend produces deterministic QoR for a given configuration (\Cref{sec:setup}), run-to-run variation arises only from the LLM-guided optimization at temperature 0.6. \Cref{tab:stability} reports the best in-loop \wns{} over three independent runs; metric units follow \Cref{tab:qor-final}.

AgenticPD achieves better mean in-loop \wns{} than the default across all six design--technology pairs, while maintaining low run-to-run variation, with standard deviations of only 0.051\,ns and 1.92\,ps for AES on SKY130HD and ASAP7, respectively.

To isolate the Judge's contribution, we replace it with a deterministic max-bottleneck heuristic that branches from the best node and restarts at the stage with the largest upstream-to-downstream \wns{} degradation. The Stage Agents, prompts, action space, and protocol remain unchanged. As shown in \Cref{tab:judge}, across three runs, the AgenticPD Judge outperforms the heuristic by 10.6\,ps \wns{} on AES and 35.5\,ps on JPEG (where the heuristic never escapes the default configuration), without material tradeoffs in the other PPA metrics: area and power differ by less than 1\% between the two judges.

This demonstrates the benefit of history-aware branch selection over greedy local bottleneck repair. Together, the proxy study establishes feedback fidelity, the branch analysis quantifies stage-aware execution, and the Judge ablation isolates the benefit of history-aware navigation.
Having established the reliability and signoff fidelity of the in-loop optimization process, all subsequent comparisons report complete post-route QoR.
\begin{figure}[t]
  \centering
  \includegraphics[width=\linewidth]{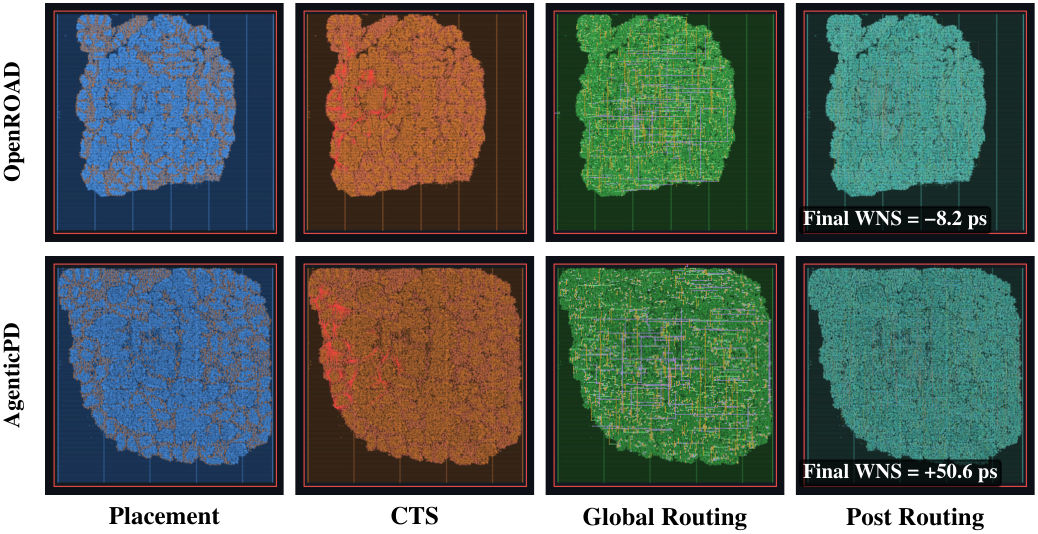}
  \caption{Layout comparison for the JPEG design on ASAP7. AgenticPD improves post-route WNS from $-8.2$\,ps to $+50.6$\,ps.}
  \label{fig:case-study-grid}
\end{figure}

\subsection{AgenticPD vs.\ Previous Tuners}
\label{sec:results}
We next compare AgenticPD against previous state-of-the-art tuners that can be applied to PD optimization problems: AutoTuner~\cite{2021ICCAD_Andrew_METRICS2.1_AutoTuner} as the representative black-box tuner and ORFS-Agent~\cite{2025MLCAD_Andrew_ORFS-Agent} as a representative LLM-based tuner. Both AgenticPD and ORFS-Agent use DeepSeek-V3.2 as the LLM engine. The OpenROAD baseline uses OpenROAD~\cite{OpenROAD} with the optimization flow provided by OpenROAD-flow-scripts (ORFS)~\cite{openroadflow}.

As shown in \Cref{tab:qor-final}, AgenticPD achieves the best \wns{} on all six benchmarks. On JPEG ASAP7, AgenticPD achieves timing closure (\wns{} = +50.6\,ps) where AutoTuner fails (\wns{} = $-$10.0\,ps) and ORFS-Agent is marginal (\wns{} = +0.3\,ps). On AES ASAP7, the hardest timing case, AgenticPD improves \wns{} by 11.1\,ps over AutoTuner and 5.2\,ps over ORFS-Agent. AgenticPD does not trade physical quality for timing: where it wins on \wns{}, it remains competitive or better in area and power.
AutoTuner~\cite{2021ICCAD_Andrew_METRICS2.1_AutoTuner} and ORFS-Agent~\cite{2025MLCAD_Andrew_ORFS-Agent} both treat the parameter space as flat $\Theta_{\mathrm{PD}}$ and restart the full flow every trial, unable to diagnose which stage causes a regression or preserve upstream progress. AgenticPD instead diagnoses the bottleneck stage and lets each Stage Agent reason within its own~$\Theta_s$ with stage-level feedback. This is not LLM prompt engineering applied to tuning; it is a purpose-built agentic system whose architecture encodes the structure of the PD optimization problem.

\subsection{Case Study: JPEG on ASAP7}
\label{sec:case-study}

\Cref{fig:case-study-grid} compares the OpenROAD layout with the AgenticPD result on JPEG (ASAP7). To illustrate how the multi-agent system reaches this result, we trace its optimization trajectory.

The Judge first directs exploration toward high core utilization: the FP Agent sets \texttt{CU}\,=\,45 and the CTS Agent selects an aggressive clock-tree configuration (\texttt{CTS\_\allowbreak CLUSTER\_\allowbreak SIZE}\,=\,60, \texttt{SLACK\_MARGIN}\,=\,0.15, \texttt{TNS\%}\,=\,10), establishing a strong floorplan-placement foundation. A later floorplan branch lowers utilization to \texttt{CU}\,=\,44, but this rebuilds the entire downstream pipeline and regresses timing---revealing that even a 1\% utilization change can disrupt a well-optimized placement--CTS configuration. The Judge reflects on this regression and shifts strategy: instead of re-exploring floorplan, it branches from the earlier node at the CTS stage, preserving the proven FP and PL stages, and the CTS Agent retunes to a balanced setting (\texttt{CTS\_\allowbreak CLUSTER\_\allowbreak SIZE}\,=\,45, \texttt{SLACK\_MARGIN}\,=\,0.13, \texttt{TNS\%}\,=\,35), reaching the selected design whose reported post-route WNS is $+50.6$\,ps without re-running any upstream stage. This trajectory instantiates the stage-aware closed loop: the measured regression changes the next branch point, while the Stage Agents retune parameters within their scopes.

%% file: figs/fig_branch_stages.tex
\definecolor{clrFP}{HTML}{7B68AE}   %
\definecolor{clrPL}{HTML}{5B9BD5}   %
\definecolor{clrCTS}{HTML}{E8915A}  %
\definecolor{clrGRT}{HTML}{70AD47}  %

\begin{figure}[t]
  \centering
  \begin{tikzpicture}[baseline={(current bounding box.center)}]
  \begin{axis}[
    xbar stacked,
    width=0.68\columnwidth,
    height=4.2cm,
    bar width=6pt,
    xlabel={},
    xmin=0, xmax=105,
    xtick={0,20,40,60,80,100},
    xticklabel style={font=\scriptsize},
    ytick=data,
    yticklabels={JPEG ASAP, JPEG SKY, IBEX ASAP, IBEX SKY,
                 AES ASAP, AES SKY},
    yticklabel style={font=\scriptsize},
    enlarge y limits=0.15,
    legend style={draw=none},
    grid=major,
    major grid style={line width=0.3pt, draw=gray!30},
    every axis plot/.append style={draw=none},
  ]
  \addplot[fill=clrFP, fill opacity=0.75] coordinates
    {(26.3,0) (15.8,1) (42.1,2) (42.1,3) (26.3,4) (15.8,5)};
  \addplot[fill=clrPL, fill opacity=0.75] coordinates
    {(15.8,0) (21.1,1) (15.8,2) (10.5,3) (36.8,4) (26.3,5)};
  \addplot[fill=clrCTS, fill opacity=0.75] coordinates
    {(31.6,0) (36.8,1) (21.1,2) (26.3,3) (15.8,4) (31.6,5)};
  \addplot[fill=clrGRT, fill opacity=0.75] coordinates
    {(26.3,0) (26.3,1) (21.1,2) (21.1,3) (21.1,4) (26.3,5)};
  \end{axis}
  \end{tikzpicture}%
  \hfill
  \begin{tikzpicture}[baseline={(current bounding box.center)}]
    \def\radius{1.1}
    \fill[clrFP, fill opacity=0.75] (0,0) -- (90:\radius) arc (90:{90-101.05}:\radius) -- cycle;
    \fill[clrPL, fill opacity=0.75] (0,0) -- ({90-101.05}:\radius) arc ({90-101.05}:{90-101.05-75.79}:\radius) -- cycle;
    \fill[clrCTS, fill opacity=0.75] (0,0) -- ({90-101.05-75.79}:\radius) arc ({90-101.05-75.79}:{90-101.05-75.79-97.89}:\radius) -- cycle;
    \fill[clrGRT, fill opacity=0.75] (0,0) -- ({90-101.05-75.79-97.89}:\radius) arc ({90-101.05-75.79-97.89}:{90-360}:\radius) -- cycle;
    \node[font=\fontsize{9}{11}\selectfont\bfseries] at ({90 - 101.05/2}:0.65)  {28.1\%};
    \node[font=\fontsize{9}{11}\selectfont\bfseries] at ({90 - 101.05 - 75.79/2}:0.65)  {21.1\%};
    \node[font=\fontsize{9}{11}\selectfont\bfseries] at ({90 - 101.05 - 75.79 - 97.89/2}:0.65)  {27.2\%};
    \node[font=\fontsize{9}{11}\selectfont\bfseries] at ({90 - 101.05 - 75.79 - 97.89 - 85.26/2}:0.65)  {23.7\%};
  \end{tikzpicture}

  \vspace{4pt}
  \begin{tikzpicture}
    \fill[black, fill opacity=0.06, rounded corners=2pt]
      (-0.15,-0.1) rectangle (8.55,0.4);
    \fill[clrFP, fill opacity=0.75]  (0,0) rectangle (0.3,0.3);
    \node[right, font=\scriptsize] at (0.35,0.15) {Floorplan};
    \fill[clrPL, fill opacity=0.75]  (1.8,0) rectangle (2.1,0.3);
    \node[right, font=\scriptsize] at (2.15,0.15) {Placement};
    \fill[clrCTS, fill opacity=0.75] (3.8,0) rectangle (4.1,0.3);
    \node[right, font=\scriptsize] at (4.15,0.15) {CTS};
    \fill[clrGRT, fill opacity=0.75] (5.4,0) rectangle (5.7,0.3);
    \node[right, font=\scriptsize] at (5.75,0.15) {Global Routing (GR)};
  \end{tikzpicture}
  \caption{Distribution of branching stages across the six benchmark optimization runs.
    Left: per-benchmark breakdown.
    Right: aggregate distribution over all benchmarks.}
  \label{fig:branch-stages}
\end{figure}

%% file: tables/tab-qor-final.tex
\begin{table*}[t]
\centering
\caption{Post-route PPA comparison across six benchmarks.
Bold\,=\,AgenticPD is the best.}
\label{tab:qor-final}
\setlength{\tabcolsep}{2pt}
\scriptsize
\resizebox{\textwidth}{!}{%
\begin{threeparttable}
\begin{tabular}{ll rrrrr rrrrr rrrrr rrrrr}
\toprule
& & \multicolumn{5}{c}{\textbf{OpenROAD~\cite{OpenROAD,openroadflow}}}
  & \multicolumn{5}{c}{\textbf{AutoTuner~\cite{2021ICCAD_Andrew_METRICS2.1_AutoTuner}}}
  & \multicolumn{5}{c}{\textbf{ORFS-Agent~\cite{2025MLCAD_Andrew_ORFS-Agent}}}
  & \multicolumn{5}{c}{\textbf{AgenticPD (Ours)}} \\
\cmidrule(lr){3-7} \cmidrule(lr){8-12} \cmidrule(lr){13-17} \cmidrule(lr){18-22}
\textbf{Tech} & \textbf{Design}
 & WNS$\uparrow$ & TNS$\uparrow$ & Area$\downarrow$ & Power$\downarrow$ & Eff.\ Freq$\uparrow$
 & WNS$\uparrow$ & TNS$\uparrow$ & Area$\downarrow$ & Power$\downarrow$ & Eff.\ Freq$\uparrow$
 & WNS$\uparrow$ & TNS$\uparrow$ & Area$\downarrow$ & Power$\downarrow$ & Eff.\ Freq$\uparrow$
 & WNS$\uparrow$ & TNS$\uparrow$ & Area$\downarrow$ & Power$\downarrow$ & Eff.\ Freq$\uparrow$ \\
\midrule
 & AES  & $-$0.321 & $-$2.2 & 125.2 & 0.424 & 207.4
        & 0.133 & 0 & 123.7 & 0.424 & 229.0
        & 0.120 & 0 & 126.8 & 0.432 & 228.3
        & \textbf{0.271} & 0 & 127.8 & 0.437 & \textbf{236.5} \\
SKY130HD
 & ibex & $-$0.664 & $-$23.8 & 181.0 & 0.110 & 93.8
        & $-$0.719 & $-$56.9 & 183.3 & 0.112 & 93.3
        & $-$0.581 & $-$22.2 & 181.6 & 0.110 & 94.5
        & \textbf{$-$0.503} & \textbf{$-$19.3} & \textbf{179.4} & \textbf{0.107} & \textbf{95.2} \\
 & JPEG & $-$0.218 & $-$2.8 & 516.3 & 0.542 & 174.9
        & $-$0.207 & $-$4.0 & 511.1 & 0.546 & 175.2
        & $-$0.202 & $-$3.6 & 519.6 & 0.549 & 175.4
        & \textbf{$-$0.154} & \textbf{$-$1.6} & 517.7 & \textbf{0.531} & \textbf{176.9} \\
\midrule
 & AES  & $-$37.614 & $-$2395.2 & 2.1 & 0.157 & 2.285
        & $-$43.399 & $-$2392.4 & 2.0 & 0.157 & 2.255
        & $-$37.444 & $-$2566.4 & 2.1 & 0.158 & 2.286
        & \textbf{$-$32.271} & \textbf{$-$1811.3} & \textbf{2.0} & \textbf{0.157} & \textbf{2.313} \\
ASAP7
 & ibex & $-$62.655 & $-$2740.9 & 2.7 & 0.047 & 0.756
        & $-$61.852 & $-$3145.1 & 2.7 & 0.048 & 0.757
        & $-$63.669 & $-$1057.6 & 2.7 & 0.047 & 0.755
        & \textbf{$-$53.867} & $-$1098.7 & 2.7 & 0.048 & \textbf{0.761} \\
 & JPEG & $-$8.229 & $-$37.4 & 6.7 & 0.092 & 1.101
        & $-$9.996 & $-$25.9 & 6.7 & 0.092 & 1.099
        & 0.337 & 0 & 6.7 & 0.092 & 1.112
        & \textbf{50.569} & \textbf{0} & 6.7 & 0.092 & \textbf{1.177} \\
\bottomrule
\end{tabular}
\begin{tablenotes}[flushleft]
\footnotesize
\item[*] WNS and TNS are reported in picoseconds for ASAP7 and nanoseconds for SKY130HD, where higher (less negative) is better ($\uparrow$). Area is in $10^3\,\mu\text{m}^2$ and power in watts, where lower is better ($\downarrow$). Eff.\ Freq is the effective frequency $1/(T_{\mathrm{clk}}-\mathrm{WNS})$~\cite{2020DAC_sklim_TPGNN}, in MHz for SKY130HD and GHz for ASAP7.
\end{tablenotes}
\end{threeparttable}%
}
\end{table*}

%% file: tables/tab-stability.tex
\begin{table}[t]
\centering
\caption{Run-to-run stability with DeepSeek-V3.2.}
\label{tab:stability}
\scriptsize
\begin{tabular}{ll r rrr r}
\toprule
\textbf{Tech} & \textbf{Design} & \textbf{Default}
 & \textbf{Run 1} & \textbf{Run 2} & \textbf{Run 3}
 & \textbf{Mean\,$\pm$\,Std} \\
\midrule
 & AES  & $-$0.526 & $+$0.234 & $+$0.144 & $+$0.232 & $+$0.203\,$\pm$\,0.051 \\
SKY130HD
 & ibex & $-$0.714 & $-$0.516 & $-$0.470 & $-$0.501 & $-$0.496\,$\pm$\,0.024 \\
 & JPEG & $-$0.386 & $-$0.363 & $-$0.386 & $-$0.386 & $-$0.378\,$\pm$\,0.013 \\
\midrule
 & AES  & $-$35.31 & $-$23.31 & $-$23.58 & $-$26.77 & $-$24.55\,$\pm$\,1.92 \\
ASAP7
 & ibex & $+$1.33  & $+$4.61  & $+$4.61  & $+$4.61  & $+$4.61\,$\pm$\,0.00 \\
 & JPEG & $+$52.21 & $+$86.35 & $+$88.45 & $+$88.45 & $+$87.75\,$\pm$\,1.21 \\
\bottomrule
\end{tabular}
\end{table}

%% file: tables/tab-judge.tex
\begin{table}[t]
\centering
\caption{Judge ablation on ASAP7.}
\label{tab:judge}
\scriptsize
\begin{tabular}{ll rrrr}
\toprule
\textbf{Design} & \textbf{Judge} & \textbf{WNS (ps)}$\uparrow$
 & \textbf{TNS (ps)}$\uparrow$ & \textbf{Area}$\downarrow$
 & \textbf{Power}$\downarrow$ \\
\midrule
AES  & Heuristic & $-$35.20\,$\pm$\,0.02 & $-$1992\,$\pm$\,140
     & 2.031\,$\pm$\,0.001 & 0.154\,$\pm$\,0.000 \\
     & AgenticPD & \textbf{$-$24.55}\,$\pm$\,1.92 & \textbf{$-$1675}\,$\pm$\,212
     & 2.044\,$\pm$\,0.013 & 0.155\,$\pm$\,0.001 \\
\midrule
JPEG & Heuristic & $+$52.21\,$\pm$\,0.00 & 0\,$\pm$\,0
     & 6.729\,$\pm$\,0.000 & 0.092\,$\pm$\,0.000 \\
     & AgenticPD & \textbf{$+$87.75}\,$\pm$\,1.21 & 0\,$\pm$\,0
     & 6.733\,$\pm$\,0.020 & 0.092\,$\pm$\,0.001 \\
\bottomrule
\end{tabular}
\end{table}

%% file: doc/5-conclusion.tex
\section{Conclusion}
\label{sec:conclusion}

In this paper, we presented AgenticPD, a stage-aware agentic framework that closes the loop between measured backend QoR and subsequent branch decisions through a reusable optimization tree.
Across vanilla LLM baselines and prior state-of-the-art tuners, AgenticPD demonstrates robust timing performance across all three LLM backbones and achieves the best timing on all six benchmarks against the latter, while remaining competitive in power and area. State reuse avoids redundant execution.
Future work will advance AgenticPD toward Agentic EDA systems that accelerate LLM-driven chip design through closed-loop agent--tool collaboration.

\section*{Acknowledgment}
This research was partially conducted by ACCESS-AI Chip Center for Emerging Smart Systems, supported by the InnoHK initiative of the Innovation and Technology Commission of the Hong Kong Special Administrative Region Government. This work is jointly supported by the Research Grants Council of Hong Kong SAR (Grant No. T46-415/25-R) and the Key Research and Development Program of Jiangsu Province (Grant No. BE2023003).

%% file: main.bbl

\begin{thebibliography}{29}


\ifx \showCODEN    \undefined \def \showCODEN     #1{\unskip}     \fi
\ifx \showISBNx    \undefined \def \showISBNx     #1{\unskip}     \fi
\ifx \showISBNxiii \undefined \def \showISBNxiii  #1{\unskip}     \fi
\ifx \showISSN     \undefined \def \showISSN      #1{\unskip}     \fi
\ifx \showLCCN     \undefined \def \showLCCN      #1{\unskip}     \fi
\ifx \shownote     \undefined \def \shownote      #1{#1}          \fi
\ifx \showarticletitle \undefined \def \showarticletitle #1{#1}   \fi
\ifx \showURL      \undefined \def \showURL       {\relax}        \fi
\providecommand\bibfield[2]{#2}
\providecommand\bibinfo[2]{#2}
\providecommand\natexlab[1]{#1}
\providecommand\showeprint[2][]{arXiv:#2}

\bibitem[Ajayi et~al\mbox{.}(2019)]%
        {OpenROAD}
\bibfield{author}{\bibinfo{person}{Tutu Ajayi}, \bibinfo{person}{Vidya~A
  Chhabria}, \bibinfo{person}{Mateus Foga{\c{c}}a}, \bibinfo{person}{Soheil
  Hashemi}, \bibinfo{person}{Abdelrahman Hosny}, \bibinfo{person}{Andrew~B
  Kahng}, \bibinfo{person}{Minsoo Kim}, \bibinfo{person}{Jeongsup Lee},
  \bibinfo{person}{Uday Mallappa}, \bibinfo{person}{Marina Neseem},
  {et~al\mbox{.}}} \bibinfo{year}{2019}\natexlab{}.
\newblock \showarticletitle{Toward an open-source digital flow: First learnings
  from the {OpenROAD} project}. In \bibinfo{booktitle}{\emph{Proceedings of the
  Design Automation Conference ({DAC})}}. \bibinfo{pages}{1--4}.
\newblock


\bibitem[{Anthropic}(2024)]%
        {Anthropic2024agents}
\bibfield{author}{\bibinfo{person}{{Anthropic}}.}
  \bibinfo{year}{2024}\natexlab{}.
\newblock \bibinfo{title}{Building Effective Agents}.
\newblock \bibinfo{howpublished}{Anthropic Research Blog}.
\newblock
\urldef\tempurl%
\url{https://www.anthropic.com/research/building-effective-agents}
\showURL{%
\tempurl}


\bibitem[{Cadence Design Systems}(2025)]%
        {Cadence_ChipStack_2025}
\bibfield{author}{\bibinfo{person}{{Cadence Design Systems}}.}
  \bibinfo{year}{2025}\natexlab{}.
\newblock \bibinfo{title}{Cadence Welcomes {ChipStack}}.
\newblock
  \bibinfo{howpublished}{\url{https://community.cadence.com/cadence_blogs_8/b/corporate-news/posts/cadence-welcomes-chipstack}}.
\newblock


\bibitem[Clark et~al\mbox{.}(2016)]%
        {technode_ASAP7}
\bibfield{author}{\bibinfo{person}{Lawrence~T Clark}, \bibinfo{person}{Vinay
  Vashishtha}, \bibinfo{person}{Lucian Shifren}, \bibinfo{person}{Aditya
  Gujja}, \bibinfo{person}{Saurabh Sinha}, \bibinfo{person}{Brian Cline},
  \bibinfo{person}{Chandarasekaran Ramamurthy}, {and} \bibinfo{person}{Greg
  Yeric}.} \bibinfo{year}{2016}\natexlab{}.
\newblock \showarticletitle{{ASAP7}: A 7-nm {finFET} predictive process design
  kit}.
\newblock \bibinfo{journal}{\emph{Microelectronics Journal}}
  \bibinfo{volume}{53} (\bibinfo{year}{2016}), \bibinfo{pages}{105--115}.
\newblock


\bibitem[{DeepSeek}(2025)]%
        {deepseek_v32_2025}
\bibfield{author}{\bibinfo{person}{{DeepSeek}}.}
  \bibinfo{year}{2025}\natexlab{}.
\newblock \bibinfo{booktitle}{\emph{{DeepSeek-V3.2} Release}}.
\newblock
\urldef\tempurl%
\url{https://api-docs.deepseek.com/news/news251201}
\showURL{%
\tempurl}


\bibitem[Fu et~al\mbox{.}(2026)]%
        {fu2026mappingevolve}
\bibfield{author}{\bibinfo{person}{Rongliang Fu}, \bibinfo{person}{Yi Liu},
  \bibinfo{person}{Qiang Xu}, {and} \bibinfo{person}{Tsung-Yi Ho}.}
  \bibinfo{year}{2026}\natexlab{}.
\newblock \showarticletitle{{MappingEvolve}: {LLM}-Driven Code Evolution for
  Technology Mapping}. In \bibinfo{booktitle}{\emph{Proceedings of the ACM/IEEE
  Design Automation Conference ({DAC})}}. \bibinfo{pages}{1--7}.
\newblock
\href{https://doi.org/10.1145/3770743.3803988}{doi:\nolinkurl{10.1145/3770743.3803988}}


\bibitem[Geng et~al\mbox{.}(2022a)]%
        {2022TCAD_PD_tuning_BO_beiyu}
\bibfield{author}{\bibinfo{person}{Hao Geng}, \bibinfo{person}{Tinghuan Chen},
  \bibinfo{person}{Yuzhe Ma}, \bibinfo{person}{Binwu Zhu}, {and}
  \bibinfo{person}{Bei Yu}.} \bibinfo{year}{2022}\natexlab{a}.
\newblock \showarticletitle{{PTPT}: Physical design tool parameter tuning via
  multi-objective Bayesian optimization}.
\newblock \bibinfo{journal}{\emph{{IEEE} Transactions on Computer-Aided Design
  of Integrated Circuits and Systems}} \bibinfo{volume}{42},
  \bibinfo{number}{1} (\bibinfo{year}{2022}), \bibinfo{pages}{178--189}.
\newblock


\bibitem[Geng et~al\mbox{.}(2022b)]%
        {2022DAC_PPATuner_beiyu}
\bibfield{author}{\bibinfo{person}{Hao Geng}, \bibinfo{person}{Qi Xu},
  \bibinfo{person}{Tsung-Yi Ho}, {and} \bibinfo{person}{Bei Yu}.}
  \bibinfo{year}{2022}\natexlab{b}.
\newblock \showarticletitle{{PPATuner}: Pareto-driven tool parameter
  auto-tuning in physical design via Gaussian process transfer learning}. In
  \bibinfo{booktitle}{\emph{Proceedings of the ACM/IEEE Design Automation
  Conference ({DAC})}}. \bibinfo{pages}{1237--1242}.
\newblock


\bibitem[Ghose et~al\mbox{.}(2025)]%
        {2025MLCAD_Andrew_ORFS-Agent}
\bibfield{author}{\bibinfo{person}{Amur Ghose}, \bibinfo{person}{Andrew~B
  Kahng}, \bibinfo{person}{Sayak Kundu}, {and} \bibinfo{person}{Zhiang Wang}.}
  \bibinfo{year}{2025}\natexlab{}.
\newblock \showarticletitle{{ORFS-Agent}: Tool-using agents for chip design
  optimization}. In \bibinfo{booktitle}{\emph{Proceedings of the ACM/IEEE
  Symposium on Machine Learning for CAD ({MLCAD})}}. IEEE,
  \bibinfo{pages}{1--13}.
\newblock


\bibitem[{Google} and {SkyWater Technology}(2020)]%
        {technode_sky130pdk}
\bibfield{author}{\bibinfo{person}{{Google}} {and} \bibinfo{person}{{SkyWater
  Technology}}.} \bibinfo{year}{2020}\natexlab{}.
\newblock \bibinfo{title}{{SkyWater SKY130} Open Source {PDK}}.
\newblock \bibinfo{howpublished}{\url{https://github.com/google/skywater-pdk}}.
\newblock
\newblock
\shownote{Accessed: 2026-03-20}.


\bibitem[Han et~al\mbox{.}(2026)]%
        {han2026leafcell}
\bibfield{author}{\bibinfo{person}{Yaohui Han}, \bibinfo{person}{Rongliang Fu},
  \bibinfo{person}{Yanming Liu}, \bibinfo{person}{Shuo Ren},
  \bibinfo{person}{Shuai Dong}, \bibinfo{person}{Yunpeng Wang},
  \bibinfo{person}{Tinghuan Chen}, \bibinfo{person}{Bei Yu}, {and}
  \bibinfo{person}{Tsung-Yi Ho}.} \bibinfo{year}{2026}\natexlab{}.
\newblock \showarticletitle{Expert-level Leaf Cell Layout Generation via
  Preference-Optimized {LLM}}. In \bibinfo{booktitle}{\emph{Proceedings of the
  International Conference on Machine Learning ({ICML})}}.
\newblock
\urldef\tempurl%
\url{https://openreview.net/forum?id=z834t47Lr4}
\showURL{%
\tempurl}


\bibitem[Hsiao et~al\mbox{.}(2024a)]%
        {2024ISPD_sklim_FastTuner}
\bibfield{author}{\bibinfo{person}{Hao-Hsiang Hsiao}, \bibinfo{person}{Yi-Chen
  Lu}, \bibinfo{person}{Pruek Vanna-Iampikul}, {and} \bibinfo{person}{Sung~Kyu
  Lim}.} \bibinfo{year}{2024}\natexlab{a}.
\newblock \showarticletitle{{FastTuner}: Transferable physical design parameter
  optimization using fast reinforcement learning}. In
  \bibinfo{booktitle}{\emph{Proceedings of the International Symposium on
  Physical Design ({ISPD})}}. \bibinfo{pages}{93--101}.
\newblock


\bibitem[Hsiao et~al\mbox{.}(2024b)]%
        {2024DAC_PD_opt}
\bibfield{author}{\bibinfo{person}{Hao-Hsiang Hsiao}, \bibinfo{person}{Pruek
  Vanna-Iampikul}, \bibinfo{person}{Yi-Chen Lu}, {and}
  \bibinfo{person}{Sung~Kyu Lim}.} \bibinfo{year}{2024}\natexlab{b}.
\newblock \showarticletitle{{ML}-based Physical Design Parameter Optimization
  for {3D ICs}: From Parameter Selection to Optimization}. In
  \bibinfo{booktitle}{\emph{Proceedings of the ACM/IEEE Design Automation
  Conference ({DAC})}}. Article \bibinfo{articleno}{252},
  \bibinfo{numpages}{6}~pages.
\newblock
\showISBNx{9798400706011}
\href{https://doi.org/10.1145/3649329.3656509}{doi:\nolinkurl{10.1145/3649329.3656509}}


\bibitem[Jung et~al\mbox{.}(2021)]%
        {2021ICCAD_Andrew_METRICS2.1_AutoTuner}
\bibfield{author}{\bibinfo{person}{Jinwook Jung}, \bibinfo{person}{Andrew~B
  Kahng}, \bibinfo{person}{Seungwon Kim}, {and} \bibinfo{person}{Ravi
  Varadarajan}.} \bibinfo{year}{2021}\natexlab{}.
\newblock \showarticletitle{{METRICS2.1} and flow tuning in the {IEEE CEDA}
  robust design flow and {OpenROAD ICCAD} special session paper}. In
  \bibinfo{booktitle}{\emph{Proceedings of the IEEE/ACM International
  Conference on Computer-Aided Design ({ICCAD})}}. IEEE, \bibinfo{pages}{1--9}.
\newblock


\bibitem[Kahng et~al\mbox{.}(2011)]%
        {2011_Andrew_PD_book}
\bibfield{author}{\bibinfo{person}{Andrew~B Kahng}, \bibinfo{person}{Jens
  Lienig}, \bibinfo{person}{Igor~L Markov}, {and} \bibinfo{person}{Jin Hu}.}
  \bibinfo{year}{2011}\natexlab{}.
\newblock \bibinfo{booktitle}{\emph{{VLSI} physical design: from graph
  partitioning to timing closure}}. Vol.~\bibinfo{volume}{312}.
\newblock \bibinfo{publisher}{Springer}.
\newblock


\bibitem[Lu et~al\mbox{.}(2020)]%
        {2020DAC_sklim_TPGNN}
\bibfield{author}{\bibinfo{person}{Yi-Chen Lu}, \bibinfo{person}{Sai
  Surya~Kiran Pentapati}, \bibinfo{person}{Lingjun Zhu},
  \bibinfo{person}{Kambiz Samadi}, {and} \bibinfo{person}{Sung~Kyu Lim}.}
  \bibinfo{year}{2020}\natexlab{}.
\newblock \showarticletitle{{TP-GNN}: A graph neural network framework for tier
  partitioning in monolithic {3D} {ICs}}. In
  \bibinfo{booktitle}{\emph{Proceedings of the 57th ACM/IEEE Design Automation
  Conference ({DAC})}}. \bibinfo{pages}{1--6}.
\newblock


\bibitem[{Moonshot AI}(2026)]%
        {kimi_k25_2026}
\bibfield{author}{\bibinfo{person}{{Moonshot AI}}.}
  \bibinfo{year}{2026}\natexlab{}.
\newblock \bibinfo{booktitle}{\emph{{Kimi-K2.5} Model by Moonshotai | {NVIDIA
  NIM}}}.
\newblock
\urldef\tempurl%
\url{https://build.nvidia.com/moonshotai/kimi-k2.5/modelcard}
\showURL{%
\tempurl}


\bibitem[{OpenAI}(2026)]%
        {Openai_chatgpt}
\bibfield{author}{\bibinfo{person}{{OpenAI}}.} \bibinfo{year}{2026}\natexlab{}.
\newblock \bibinfo{booktitle}{\emph{{ChatGPT}}}.
\newblock
\urldef\tempurl%
\url{https://openai.com/research}
\showURL{%
\tempurl}


\bibitem[OpenCores(2002a)]%
        {design_aes}
\bibfield{author}{\bibinfo{person}{OpenCores}.}
  \bibinfo{year}{2002}\natexlab{a}.
\newblock \bibinfo{title}{{AES} (Rijndael) {IP} Core}.
\newblock
  \bibinfo{howpublished}{\url{https://opencores.org/projects/aes_core}}.
\newblock
\newblock
\shownote{Accessed: 2026-03-20}.


\bibitem[OpenCores(2002b)]%
        {design_jpeg}
\bibfield{author}{\bibinfo{person}{OpenCores}.}
  \bibinfo{year}{2002}\natexlab{b}.
\newblock \bibinfo{title}{{JPEG} Encoder, Video Compression Systems Project}.
\newblock
  \bibinfo{howpublished}{\url{https://opencores.org/projects/video_systems}}.
\newblock
\newblock
\shownote{Accessed: 2026-03-20}.


\bibitem[{Qwen Team}(2026)]%
        {qwen35_2026}
\bibfield{author}{\bibinfo{person}{{Qwen Team}}.}
  \bibinfo{year}{2026}\natexlab{}.
\newblock \bibinfo{booktitle}{\emph{{Qwen3.5}: Towards Native Multimodal
  Agents}}.
\newblock
\urldef\tempurl%
\url{https://qwen.ai/blog?id=qwen3.5}
\showURL{%
\tempurl}


\bibitem[Schiavone et~al\mbox{.}(2017)]%
        {design_ibex}
\bibfield{author}{\bibinfo{person}{Pasquale~Davide Schiavone},
  \bibinfo{person}{Francesco Conti}, \bibinfo{person}{Davide Rossi},
  \bibinfo{person}{Michael Gautschi}, \bibinfo{person}{Antonio Pullini},
  \bibinfo{person}{Eric Flamand}, {and} \bibinfo{person}{Luca Benini}.}
  \bibinfo{year}{2017}\natexlab{}.
\newblock \showarticletitle{Slow and steady wins the race? A comparison of
  ultra-low-power {RISC-V} cores for Internet-of-Things applications}. In
  \bibinfo{booktitle}{\emph{Proceedings of the International Symposium on Power
  and Timing Modeling, Optimization and Simulation ({PATMOS})}}. IEEE,
  \bibinfo{pages}{1--8}.
\newblock


\bibitem[Sharma et~al\mbox{.}(2024)]%
        {2024MLCAD_OpenROAD_Assistant}
\bibfield{author}{\bibinfo{person}{Utsav Sharma}, \bibinfo{person}{Bing-Yue
  Wu}, \bibinfo{person}{Sai Rahul~Dhanvi Kankipati}, \bibinfo{person}{Vidya~A
  Chhabria}, {and} \bibinfo{person}{Austin Rovinski}.}
  \bibinfo{year}{2024}\natexlab{}.
\newblock \showarticletitle{{OpenROAD-Assistant}: An open-source large language
  model for physical design tasks}. In \bibinfo{booktitle}{\emph{Proceedings of
  the ACM/IEEE International Symposium on Machine Learning for CAD ({MLCAD})}}.
  \bibinfo{pages}{1--7}.
\newblock


\bibitem[{Siemens Digital Industries Software}(2026)]%
        {Siemens_fuse_2026}
\bibfield{author}{\bibinfo{person}{{Siemens Digital Industries Software}}.}
  \bibinfo{year}{2026}\natexlab{}.
\newblock \bibinfo{title}{Siemens launches Fuse {EDA AI} Agent}.
\newblock
  \bibinfo{howpublished}{\url{https://news.siemens.com/en-us/siemens-fuse-eda-ai-agent/}}.
\newblock


\bibitem[{Synopsys}(2026)]%
        {Synopsys_AgentEngineer_2026}
\bibfield{author}{\bibinfo{person}{{Synopsys}}.}
  \bibinfo{year}{2026}\natexlab{}.
\newblock \bibinfo{title}{Agentic {AI}: Automating Engineering Workflows with
  {AI} Agents}.
\newblock
  \bibinfo{howpublished}{\url{https://www.synopsys.com/ai/agentic-ai.html}}.
\newblock


\bibitem[{The OpenROAD Project}(2024)]%
        {openroadflow}
\bibfield{author}{\bibinfo{person}{{The OpenROAD Project}}.}
  \bibinfo{year}{2024}\natexlab{}.
\newblock \bibinfo{title}{{OpenROAD-flow-scripts}: Autonomous Digital Design
  Flow}.
\newblock
  \bibinfo{howpublished}{\url{https://github.com/The-OpenROAD-Project/OpenROAD-flow-scripts}}.
\newblock
\newblock
\shownote{Accessed: 2026-03-20}.


\bibitem[Wu et~al\mbox{.}(2025)]%
        {2025ICLAD_OpenROAD_Agent}
\bibfield{author}{\bibinfo{person}{Bing-Yue Wu}, \bibinfo{person}{Utsav
  Sharma}, \bibinfo{person}{Austin Rovinski}, {and} \bibinfo{person}{Vidya~A
  Chhabria}.} \bibinfo{year}{2025}\natexlab{}.
\newblock \showarticletitle{{OpenROAD} Agent: An Intelligent Self-Correcting
  Script Generator for {OpenROAD}}. In \bibinfo{booktitle}{\emph{Proceedings of
  the IEEE International Conference on LLM-Aided Design ({ICLAD})}}. IEEE,
  \bibinfo{pages}{16--22}.
\newblock


\bibitem[Wu et~al\mbox{.}(2024)]%
        {2024TCAD_byu_ChatEDA}
\bibfield{author}{\bibinfo{person}{Haoyuan Wu}, \bibinfo{person}{Zhuolun He},
  \bibinfo{person}{Xinyun Zhang}, \bibinfo{person}{Xufeng Yao},
  \bibinfo{person}{Su Zheng}, \bibinfo{person}{Haisheng Zheng}, {and}
  \bibinfo{person}{Bei Yu}.} \bibinfo{year}{2024}\natexlab{}.
\newblock \showarticletitle{{ChatEDA}: A large language model powered
  autonomous agent for {EDA}}.
\newblock \bibinfo{journal}{\emph{{IEEE} Transactions on Computer-Aided Design
  of Integrated Circuits and Systems}} \bibinfo{volume}{43},
  \bibinfo{number}{10} (\bibinfo{year}{2024}), \bibinfo{pages}{3184--3197}.
\newblock


\bibitem[Zheng et~al\mbox{.}(2023)]%
        {2023TODAES_beiyu_REMOTune}
\bibfield{author}{\bibinfo{person}{Su Zheng}, \bibinfo{person}{Hao Geng},
  \bibinfo{person}{Chen Bai}, \bibinfo{person}{Bei Yu}, {and}
  \bibinfo{person}{Martin~DF Wong}.} \bibinfo{year}{2023}\natexlab{}.
\newblock \showarticletitle{Boosting {VLSI} design flow parameter tuning with
  random embedding and multi-objective trust-region Bayesian optimization}.
\newblock \bibinfo{journal}{\emph{{ACM} Transactions on Design Automation of
  Electronic Systems}} \bibinfo{volume}{28}, \bibinfo{number}{5}
  (\bibinfo{year}{2023}), \bibinfo{pages}{1--23}.
\newblock


\end{thebibliography}
